\documentclass{ifacconf}
\pdfoutput = 1

\usepackage{graphicx}      % include this line if your document contains figures
\usepackage{natbib}        % required for bibliography
\usepackage{xcolor}    %to add colours
\usepackage{amsmath} 
\usepackage{multirow}
\usepackage{amssymb}  % assumes amsmath package installed
\renewcommand\L{\mathcal{L}}
\newcommand{\p}{\partial}
\newcommand{\R}{\mathbb{R}}

\usepackage{soul} %for \st
\begin{document}
\begin{frontmatter}

\title{Discrete Lagrangian Neural Networks with Automatic Symmetry Discovery} 

\author[First]{Yana Lishkova\textsuperscript{$\dagger$}} 
\author[Second]{Paul Scherer\textsuperscript{$\dagger$}} 
\author[First]{Steffen Ridderbusch}
\author[Second]{Mateja Jamnik}
\author[Second]{Pietro Li\`o}
\author[Fourth]{Sina Ober-Bl\"obaum}
\author[Fourth]{Christian Offen}

\address[First]{University of Oxford, Oxford, UK }
\address[Second]{University of Cambridge, Cambridge CB3 0FD,  UK }
\address[Fourth]{Paderborn University, D-33098 Paderborn,
Germany}

\begin{abstract}
By one of the most fundamental principles in physics, a dynamical system will exhibit those motions which extremise an action functional. This leads to the formation of the Euler-Lagrange equations, which serve as a model of how the system will behave in time. If the dynamics exhibit additional symmetries, then the motion fulfils additional conservation laws, such as conservation of energy (time invariance), momentum (translation invariance), or angular momentum (rotational invariance). To learn a system representation, one could learn the discrete Euler-Lagrange equations, or alternatively, learn the discrete Lagrangian function $\mathcal{L}_d$ which defines them. Based on ideas from Lie group theory, we introduce a framework to learn a discrete Lagrangian along with its symmetry group from discrete observations of motions and, therefore, identify conserved quantities. The learning process does not restrict the form of the Lagrangian, does not require velocity or momentum observations or predictions and incorporates a cost term which safeguards against unwanted solutions and against potential numerical issues in forward simulations. The learnt discrete quantities are related to their continuous analogues using variational backward error analysis and numerical results demonstrate the improvement such models can have both qualitatively and quantitatively even in the presence of noise.
\end{abstract}

\begin{keyword}
Variational integrators, Neural networks, Symmetry, Nonlinear system identification%, Grey box modeling: EQUALS}
\end{keyword}

\end{frontmatter}
%===============================================================================
\def\thefootnote{\dagger}\footnotetext{These authors contributed equally to this work.}\def\thefootnote

\section{Introduction}

It is a well-established theory from the field of variational calculus that the behaviour of an unforced system can be derived from the Lagrange-d'Alembert principle using only knowledge of the system's Lagrangian. The resulting Euler-Lagrange equations of motion act as a model describing how the system's configuration and velocity evolve as time progresses from an initial displacement and thus serve as a state space system description. Equivalently, one could describe the system in the phase space through the canonical equations of motion based only on knowledge of the system's Hamiltonian. To learn a system representation, one could identify the equations of motion directly, or alternatively, identify the Hamiltonian or the Lagrangian which defines them. Using these approaches \cite{greydanus2019hamiltonian} proposed the Hamiltonian Neural Network (HNN), closely followed by the Lagrangian Neural Networks (LNN) by \cite{cranmer2020lagrangian}.  
In combination with model reduction the Lagrangian learning technique has been applied to different real-life high dimensional problems such as video prediction \citep{allenblanchette2020lagnetvip}.

Drawing from the field of mathematical physics we know that Hamiltonian and Lagrangian models encode additional information about the system such as symmetries. Through Noether's theorem, these symmetries relate to conservation laws which further restrict the motion of the system. The use of specialised integrators, known as geometric integrators, allows the preservation of these characteristics from the continuous into the discrete domain. This results in a better qualitative representation of the system and a preservation of the symmetries and energy accurately or up to small bounded oscillations for exponentially long times \citep{Reading35}.

Variational integrators are geometric integrators that are based on the theory of discrete mechanics \citep{Reading57}. With standard integrators, the equations of motion are first derived and then discretised. With variational integrators a discrete approximation of the Lagrangian is created and then used in a discrete Lagrange-d'Alembert principle leading to equations of motion which are structure-preserving. With this in mind \cite{saemundsson2020variational} proposed learning of the Lagrangian with forward simulations using the variational integrator and demonstrated accurate long-term predictions even when learning from noisy data. This was extended for systems with external forcing by \cite{havens2021forced}. However, both works rely on a phase space formulation of the variational integrator, which requires data and prediction of both the configuration and the momentum. 

When using a variational integrator, we can use a second formulation which removes the need to store observations or calculate predictions of the canonical momentum or the velocity in both the learning and testing stages. In this formulation an initial momentum value and the observations and predictions of the configuration are sufficient to model the system behaviour in time. This is an advantage compared to the use of standard integrators as noted by \cite{aoshima2021deep} for an alternative energy-preserving method in the absence of external forcing. Another important competing approach are the symplectic recurrent neural networks (SRNN) by \cite{chen2019symplectic}, which learn the Hamiltonian and rely on a different symplectic integrator that is performant even in noisy systems. However, it again relies on observations from both the configuration and the momentum.

The approaches with variational integrators mentioned so far learn a continuous Lagrangian expression from discrete data and use a discrete approximation of this expression in the forward variational simulations. These approximations introduce discretisation errors in the results which could amplify the modelling error of a learnt Lagrangian (\cite{LSI}). To prevent these errors one can instead learn the discrete Lagrangian and use it for variational simulations in testing and training as demonstrated by \cite{qin2020machine} with numerical experiments on nonlinear oscillations and the Kepler problem, by  \cite{santos2022symplectic} for an assumed structure of a potential and a kinetic energy term learned with two separate neural networks, and in a PDE setting by \cite{PDE_LDensity} for discrete Lagrangian densities.

%Despite these investigations, the question still remains: once we learn either the continuous or the discrete Lagrangian how can we best obtain one from the other? This question is non-trivial and goes back to the field of backward error analysis \citep{Reading35}.

Discrete Hamiltonians and discrete Lagrangians can be related to their continuous counterparts by backward error analysis \citep{Reading35, vermeeren2017modified}. \cite{LSI} and \cite{HSI} exploit backward error analysis in a machine learning context in for analysis purposes as well as to compensate for discretisation errors when the learned discrete Lagrangian is used in numerical simulations to compute motions to given initial values. Using Gaussian Processes (GPs) they learn a so called inverse modified (continuous) Lagrangian that, after discretisation with a variational integrator, is consistent with the motion data of the true system (see Figure 1). Their regularisation terms are, however, tailored to GPs and are not suitable for neural networks.

Our proposed methodology aims to learn an inverse modified discrete Lagrangian without prior restriction on its form, incorporating additional steps to build upon the strengths of previous approaches and subvert their limitations. For example, learning the inverse modified discrete form avoids discretisation error and requires observations only of the configuration alone, without the need to obtain or simulate the velocity. Furthermore, we propose an additional \textit{degeneracy} cost term which inductively biases the network to avoid learning discrete Lagrangians whose equations of motion have degenerate roots and pose problems for forward simulation incorporating numerical root-solving techniques. The additional term also inductively biases the neural network to avoid learning non-desirable constant Lagrangian solutions. With these improvements, the first of our two proposed methods, the \textit{Discrete Lagrangian Neural Network} (DLNN) successfully incorporates the use of variational integrators and variational backward error analysis in a novel neural network based model, which uses the underlying structure of the problem to improve the representation qualitatively and explicitly guard against numerical issues in forward integration.

As another contribution, we introduce a novel method to automatically learn variational symmetries of a dynamical system along with the system's discrete Lagrangian. More precisely, we introduce a framework that for a given Lie group action on the configuration space identifies a subgroup which acts by symmetries. Through construction of a momentum map, conserved quantities are derived. While our framework considers variational symmetries of discrete actions, a framework for symplectic symmetries of Hamiltonian systems (SymHNN) was developed in \cite{SymHNN}. Inductive biases borne out of the incorporation of symmetries into learning algorithms have previously shown desirable behaviours in the resulting models such as reduced sample complexity and improved generalisation whilst significantly reducing model complexity \citep{dehmamy2021automatic}.Our framework is explicitly worked out for the group of affine linear transformations, which consists of arbitrary compositions of translation, rotations, and scaling transformations. We demonstrate in numerical examples that our proposed \textit{Symmetric Discrete Lagrangian Neural Network} (SymDLNN) successfully learns symmetries and conservation laws and that incorporation of symmetries improves the predictions compared to previous approaches.

To summarise, \textbf{our contributions} are:
\begin{itemize}
    % \item A new regularisation term for learning discrete Lagrangians from snapshot data of motions. The regularisation guarantees that the learnt discrete Lagrangian can be efficiently employed in the numerical computation of trajectories.
    % \item 
    % A novel framework to automatically identify symmetries and conservation laws while a discrete Lagrangian is learned.
    
    % \item The use of variational backward error analysis for
    % %in the setting of 
    % learning discrete Lagrangians and symmetries.

\item {\bf Symmetry framework.} We introduce a novel framework based on Lie group theory to automatically discover and incorporate symmetries and conservation laws during model training.

    \item {\bf Numerical analysis informed learning.}
    We propose a novel methodology for learning inverse modified discrete Lagrangians from snapshot observations of motions. Our new regularisation term guarantees that the learned model can be efficiently employed in the computation of trajectories.

    \item Use of variational backward error analysis in combination with the learnt discrete Lagrangian.

\end{itemize}

%%%%%%%%%%%%%%%%%%%%%%%%%%%%%%%%%%%%%%%%%%%%%%%%%%%%%%%%%%%%%%%%%%%%%%%%%%%
\section{Background}
\subsection{Continuous Lagrangian dynamics }
Consider a mechanical system, whose configuration $q(t)\in \mathbb{R}^{n_q}$ evolves on a configuration manifold $Q$, with associated tangent bundle $TQ$. The behaviour of the system in time can be described in state space by the evolution of its configuration vector $q(t)\in Q$ and its associated velocity vector  $\dot{q}(t)$ such that $(q(t),\dot{q}(t)) \in TQ$. The system is assumed to possess a regular Lagrangian  $\L: TQ \to \mathbb{R}$, which is not explicitly time dependent. It is known that the motion of the system is governed by the Lagrange-d'Alembert principle, which requires that
\begin{equation}
	\label{eq:LdA1}
\delta \int_{t_0}^{t_f} \L(q,\dot{q})\,dt =0 \notag
\end{equation}
is satisfied for all variations $\delta q$ with $\delta q(t_0) = \delta q(t_f) = 0$ \citep{Reading57}. Using integration by parts 
this principle results in the following equations:
 \begin{equation}
     D_1 \mathcal{L}(q,\dot{q})-\frac{d}{dt}D_2 \mathcal{L}(q,\dot{q})=0
      \label{eqn:cELE}
 \end{equation}
known as the Euler-Lagrange equations of motion \citep{liberzon2011calculus}. Here $D_i$ is the partial derivative operator with respect to the $i$-th argument and the dependencies on time have been omitted to simplify the notation. Using these equations one can model how the system evolves in the state space over time. The system's evolution can equivalently be described in phase space using the configuration vector $q$ and conjugate momenta $p$ defined as $
    p = D_2\mathcal{L}(q,\dot{q})$. Notably for a given system, the Lagrangian which satisfies Equation (\ref{eqn:cELE}) is not unique and thus the conjugate momentum is not unique either. For example, any constant function could satisfy the equation but would not lead to any dynamics. This is important for schemes attempting to learn the Lagrangian function as one would need to safeguard against constant solutions.

\subsection{Discrete Lagrangian mechanics} \label{section:dEOM}
In the discrete time setting $TQ$ is replaced with $Q \times Q$ and a discrete configuration path is defined as $q_{\,d}(t_k) = q_{\,k}$ with  $q_k \approx q(t_k)$  for $ t_k = t_0+ k\Delta T \textrm{ and } k=0,\ldots,N $ where $N=t_f/\Delta t$.  Based on this discretisation a discrete Lagrangian $\mathcal{L}_d$ is an approximation
\begin{equation}\label{eq:Ld}
    \L_{d}(q_k,q_{k+1})\approx\int_{t_k}^{t_{k+1}} \L(q,\dot{q}) \, dt, 
\end{equation}
where $q$ on the right hand side of \eqref{eq:Ld} solves the boundary value problem \eqref{eqn:cELE} with $ q(t_k)=q_{k}$, $q(t_{k+1})=q_{k+1}$. A discrete version of the Lagrange-d' Alembert principle
\[
    \delta \sum\nolimits_{i=0}^{N-1} \L_d(q_k,q_{k+1}) = 0
\]
leads to the discrete Euler-Lagrange equations:
\begin{equation}
    D_2 \L_d(q_{k-1},q_k) + D_1\L_d(q_{k},q_{k+1}) = 0
    \label{eqn:dELE}
\end{equation}
To parallel the continuous theory one can define the discrete conjugate momentum $p_k$ as
$ 
     p_k =-  D_1\L_d(q_{k},q_{k+1})$
(see \cite{Reading57}). Given $q_0$ and $q_1$ one can use equations (\ref{eqn:dELE})  to time-step in time to compute $q_{k}$ for $k>1$. To compute $q_1$ and initiate the integration one can use the expression for the discrete momentum for the initial position $q_0$ and  conjugate momentum $p_0$. Similarly to the continuous section techniques learning the discrete Lagrangian would need to safeguard against solutions which would satisfy Equation (\ref{eqn:dELE}) but would not produce meaningful dynamics.
\subsection{Standard and variational backward error analysis} \label{section:BEA}
Assume we have a system with continuous dynamics %of the form   
$\dot{x}(t) = f(x(t))$
and a numerical integrator $
    \dot{x}_{k+1} = z(x_k)$.
This discrete method is only an approximation of the actual dynamics $x(k\Delta t)$ at time $t=k \Delta t$ or in other words $x_k \approx x(k \Delta t)$. Backward error analysis focuses on discovering dynamics $\dot{x}^m=f^m(x^m(t))$ for which equation $\dot{x}_{k+1} = z(x_k)$ is in fact the exact solution or in other words $x_k = x^m(k\Delta t)$. The dynamics $\dot{x}^m=f^m(x^m(t))$ are often termed as the modified equations. By studying the difference between $x_k$ and $x^m(k \Delta t)$ one can determine important properties of the integrator scheme \citep{Reading35}. 

\begin{figure} [!tbp]
    \centering
  \includegraphics[scale=0.25]{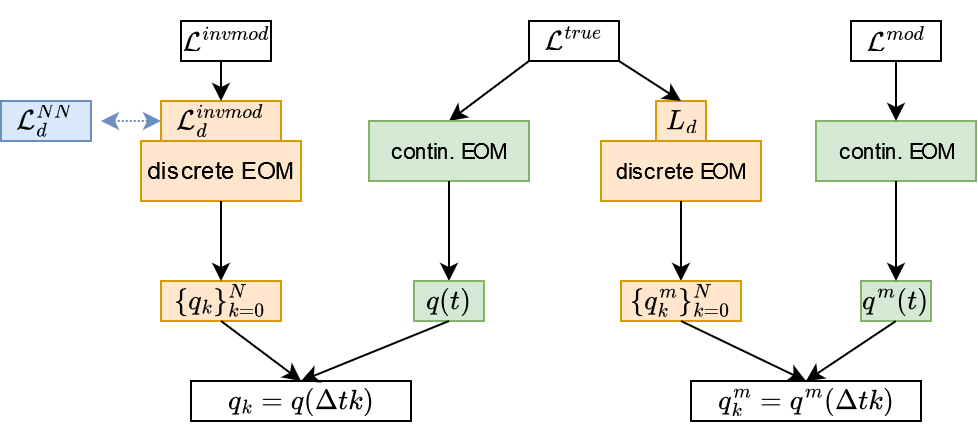}
    \caption{Diagramatic explanation of the modified and inverse modified Lagrangian functions (contin.-continuous, EOM- equations of motion).}
    \label{fig:trajplot}
\end{figure}

Following from this idea, \cite{vermeeren2017modified} has developed a similar analysis for variational integrators, known as variational backward error analysis. As depicted in Figure \ref{fig:trajplot}, the true dynamics $q(t)$ of the system  are governed by the continuous equations (\ref{eqn:cELE}) when the true system Lagrangian is used $\mathcal{L}^{true}$. The solutions of Equation (\ref{eqn:dELE}) $\{q^m_k\}_{k=0}^{N}$ are only approximations of these dynamics. On the other hand \cite{vermeeren2017modified} derived a modified Lagrangian expression $\mathcal{L}^{mod}$ which when plugged in (\ref{eqn:cELE}) would obtain a solution  $q^m(t)$ such that  $q^m(k \Delta t) = q^m_k$. \cite{LSI,HSI}  on the other hand searched for an inverse modified Lagrangian $\mathcal{L}^{invmod}$ expression which, when plugged into the variational scheme would obtain discrete path $q_k$ which accurately represents the true dynamics in the discrete domain i.e. $q_k = q(k \Delta t)$. He further describes how  $\mathcal{L}^{invmod}$ can be used to obtain  $\mathcal{L}^{true}$.

In this article, we propose to learn discrete inverse modified Lagrangians directly from data. Once it is learned, motions can be computed using Equation \eqref{eqn:dELE}. To initialise the computation, initial data consisting of the first two positions $(q_0,q_1)$ of the trajectory are required. 
%However, to solve initial value problems with initial data consisting of initial position and velocity $(q_0,\dot{q}_0)$ it is necessary to compute the continuous Lagrangian of the system from the learnt discrete Lagrangian such that variational integration can proceed. For this and for the purpose of system identification the inverse modified Lagrangian $\mathcal{L}^{invmod}$ 
 For the purpose of system identification, the inverse modified Lagrangian $\mathcal{L}^{invmod}$  can be computed from its discrete learnt counterpart by a change of variables depending on the variational integrator and subsequently using variational backward error analysis formulas used to identify a function for the true Lagrangian, which will be denoted $\mathcal{L}^{VBEA}$. 

For example, for a simulation using a step of size $\Delta t$ and the variational midpoint rule in the one-dimensional case $\mathcal{L}^{invmod}(q,\dot{q}) = \mathcal{L}_d^{invmod}(q-\Delta t/2 \dot {  q},  q+\Delta t/2 \dot {  q})$, where $\mathcal{L}_d^{invmod}$ is the discrete Lagrangian that we learn, and 
\begin{align}
	\mathcal{L}^{VBEA}  &= \mathcal{L}^{invmod}+\tfrac{1}{24} \Delta t^2\frac{(\p_q {\mathcal{L}^{invmod}}- \p^2_{q \dot{q}}{\mathcal{L}^{invmod}} \dot{q})^2}{\p^2_{\dot{q}\dot{q}}{\mathcal{L}^{invmod}}} \notag \\ & - \tfrac{1}{24} \Delta t^2\p_{qq}{\mathcal{L}^{invmod}} \dot{  q}^2 +O(\Delta t^4).  \label{eqn:MPBEA}
\end{align}
Formulas for higher dimensions or truncation orders can be derived as explained in \cite{vermeeren2017modified}.
%Notice that a learnt discrete Lagrangian can be interpreted as the discretisation of the inverse modified Lagrangians $\mathcal{L}_d^{invmod}$, not the  the discretisation of the true Lagrangian $\mathcal{L}_d^{true}$ as explained by Figure \ref{fig:trajplot}.

\subsection{Variational symmetries, conserved quantities, and a framework for our symmetry learning method}\label{sec:LieGroup}

Dynamical systems governed by Euler-Lagrange equations can exhibit symmetries. These are of great significance since they encode important qualitative features of motions: for instance, the idealised motion of a pendulum on a cart is described by Equation \eqref{eqn:cELE} with the Lagrangian
\begin{equation}\label{eqn:LCart}
\mathcal{L}_\mathrm{cp}(s,\phi,\dot{s},\dot{\phi}) = \frac 12 (\alpha \dot{\phi}^2 + 2 \beta \cos(\phi) \dot{s}\dot{\phi} + \gamma \dot{s}^2 )+D \cos(\phi),
\end{equation}
with position variables $q=(s,\phi)$ and velocity variables $\dot q=(\dot s,\dot \phi)$,
where $\alpha = m_1l^2$, $\beta = m_1l$, $\gamma =m_2+m_1$, and $D=-m_1gl$. Here $m_2$ is the mass of the cart, $m_1$ the mass of the pendulum, $l$ the length of the pendulum, $g$ a gravity acceleration constant. The Lagrangian $\mathcal{L}_\mathrm{cp}$ does not depend explicitly on the variable $s$. In other words, it is invariant under translations of $s$, that is, the action $s \mapsto s + \eta$ for $\eta \in \R$ and $\frac{\p \mathcal{L}_\mathrm{cp}(q,\dot{q})}{\p s} = 0$ for all $(q,\dot{q})$. By Noether's theorem, this symmetry corresponds to the conservation of the conjugate momentum
\begin{equation}
   I_{\mathrm{cp}} = \frac{\p {\mathcal L}_{\mathrm{cp}}}{\p \dot{s}} = \beta \cos(\phi) \dot{\phi} + \gamma \dot{s}
\end{equation}
along motions $q(t)=(s(t),\phi(t))$.
More generally, if for a Lagrangian $\mathcal{L}(q,\dot{q})$ with $q \in Q = \R^{n_d}$ there exists a direction $w \in \R^{n_d}$ such that $\mathcal{L}$ is invariant under an action $(q,\dot{q}) \mapsto (  q + s w ,\dot{  q})$ for all $s \in \R$ then the directional derivative
\begin{equation}
    w^\top \nabla_{q} \mathcal{L} (q,\dot{q})
    = \sum\nolimits_{j=1}^{n_d} w_j \frac{\p \mathcal{L}}{\p q_j} (q,\dot{q}) =0
\end{equation}
and the quantity
\begin{equation}
    I(q,\dot{q}) = w^\top \nabla_{\dot q} \mathcal{L}(q,\dot{q}) = \sum\nolimits_{j=1}^{n_d} w_j \frac{\p {\mathcal L}}{\p \dot{q}_j}(q,\dot{q})
\end{equation}
are preserved along motions $q(t)$ of Equation \eqref{eqn:cELE}. 
Even more generally, an affine linear transformation is described by an invertible matrix $W \in \mathrm{Gl}(\R,n_q)$ and a vector $w \in \R^{n_q}$ and acts on $(q,\dot{q}) \in TQ$ by $(q,\dot{q}) \mapsto (Wq+w,W\dot{q})$. Here $\mathrm{Gl}(\R,n_q)$ denotes the group of invertible matrices. The matrix $W$ encodes a rotation and scaling, and the vector $w$ encodes a translation. We can write $W$ and $w$ into a matrix $\begin{pmatrix}
W&w\\ 0 & 1
\end{pmatrix} \in \mathrm{Gl}(\R,n_q+1)$.
Affine transformations form a group $G$ which can be represented by the following subgroup of $(n_q+1) \times (n_q+1)$-dimensional invertible matrices
\[
G_{\mathrm{aff}} = \left\{ \left.\begin{pmatrix}
W&w\\ 0 & 1
\end{pmatrix} \, \right| \, A \in \mathrm{Gl}(\R,n_q), \, w \in \R^{n_q} \right\}.
\]
Any 1-dimensional subgroup of $G_{\mathrm{aff}}$ can be defined by a matrix $M \in \R^{n_q \times n_q}$ and vector $w \in \R^{n_q}$ and is of the form
\[G_{(M,w)} = \left\{
\left.\exp\left(\begin{pmatrix}
\eta M& \eta w\\ 0 & 1
\end{pmatrix} \right) \right| \eta \in \R \right \} \subset G,
\]
where $\exp$ is the matrix exponential.
If $G_{(M,w)}$ acts by symmetries, that is, if the Lagrangian $\mathcal L$ is invariant under actions by the elements in $G_{(M,w)}$, then 
\begin{equation}\label{eq:checkAffine}
    (Mq+w)^\top \nabla_{q} \mathcal{L} (q,\dot{q})
    + M^\top \nabla_{\dot{q}} \mathcal{L} (q,\dot{q})
    =0
\end{equation}
for all $(q,\dot{q}) \in TQ$ and the quantity
\begin{equation}\label{eq:AffConserved}
    I(q,\dot{q}) = (M  q+w)^\top \nabla_{\dot{q}} \mathcal{L}(q,\dot{q})
\end{equation}
is conserved along motions defined by the Euler-Lagrange Equation \eqref{eqn:cELE}.
When working with discrete Lagrangians $\mathcal{L}_d \colon Q \times Q \to \R$, the symmetry condition \eqref{eq:checkAffine} is replaced by
\begin{equation}\label{eq:checkAffineLd}
     (M  q_0+w)^\top \nabla_{q_0} \mathcal{L} ( q_0, q_1)
    +(M q_1+w)^\top \nabla_{q_1} \mathcal{L} ( q_0, q_1)
    =0
\end{equation}
for all $(q_0,q_1) \in Q \times Q$ and the conserved quantity for discrete motions becomes 
\begin{equation}\label{eq:AffConservedLd}
    I( q_k, q_{k+1}) = -(M q_k+w)^\top \nabla_{q_k} \mathcal{L}( q_k, q_{k+1}).
\end{equation}

For details, we refer to the book by \cite{MarsdenRatiu99}.
Our method SymDLNN automatically identifies subgroups $G_{(M,w)}$ of the group of affine linear transformations under which a (discrete) Lagrangian is invariant while the (discrete) Lagrangian of a system is learned. As a consequence, we identify the corresponding conserved quantity as well.

Lagrangians are not uniquely determined by the system's motion and non-symmetric Lagrangians can govern highly symmetric dynamical systems. Our method will guide the learning process to a symmetric Lagrangian. This regularises the learning process and improves predictions as we will demonstrate in numerical examples. Our approach is not restricted to affine linear symmetries or 1-dimensional symmetry groups. To present the most general framework, let us briefly introduce Lie group actions and invariant vector fields. See \cite{MarsdenRatiu99} for details.
For a Lie group $G$, let $\mathfrak{g}$ denote its Lie algebra and $\exp \colon \mathfrak{g} \to G$ the exponential map. Consider a group action $a \colon G \to \mathrm{Diff}(Q), \, g \mapsto a_g$.
Here  $\mathrm{Diff}(Q)$  denotes the group of diffeomorphisms on $Q=\R^{n_q}$. In the setting of continuous Lagrangians, the group action can be prolonged to an action 
$A \colon G \to \mathrm{Diff}(TQ)$, $g \mapsto A_g$ by defining $A_g(q,\dot{q})=(a_g( q ),D a_g( q) \dot{ q})$, 
where $D a_g( q)$ is the Jacobi
matrix of the diffeomorphism $a_g$ at $q$. Here we have identified $\mathcal{M}=TQ \cong \R^{n_q} \times \R^{n_q}$. For the setting of discrete Lagrangians, the diagonal group action $A \colon G \to \mathrm{Diff}(Q\times Q)$, $g \mapsto A_g$ with $A_g(q_0,q_1)=(a_g(q_0),a_g(q_1))$ is considered instead of the prolonged group action and $\mathcal M = Q \times Q$.
In both cases, for $v \in \mathfrak{g}$ the left invariant vector field $\hat{v} \in \mathfrak{X}(\mathcal{M})$ is defined by
\[
\hat{v}_z = \left.\frac{\d}{\d t}\right|_{t=0} A_{\exp{(tv)}}(z) \in T_z \mathcal{M}, \quad z \in \mathcal{M}.\]
These vector fields can be thought of as infinitesimal actions of the Lie group $G$ on $\mathcal{M}$.

The idea of SymDLNN is to identify a basis $v^1,\ldots,v^K$ of a subspace of $V^{(K)} \subset \mathfrak{g}$ such that $\hat{v}^j_z(\mathcal L) =0$ for all $z \in \mathcal{M}$ and $j=1,\ldots, K$. Under mild assumptions on the group action, a momentum map $J \colon M \to V^{(K)\ast}$ can be constructed from which $K$ functionally independent conserved quantities can be computed as $I^j(q,\dot{q}) = \langle J(q,\dot{q}) , v^j \rangle$, where $\langle \cdot, \cdot \rangle$ denotes the dual pairing.
Relating this theory back to the example of affine linear symmetries, the Lie algebra for the affine linear group $G_\mathrm{aff}$ can be written as the following subspace of $\R^{(n_q+1) \times(n_q+1)}$:
\begin{equation}\label{eq:gAffSym}
\mathfrak{g}= \left\{ \left.\begin{pmatrix}
	M&w\\ 0 & 0
\end{pmatrix} \, \right| \, M \in \R^{n_q\times n_q}, \, w \in \R^{n_q} \right\}
\end{equation}
SymDLNN seeks as many elements $v^j = \begin{pmatrix}
	M^j&w^j\\ 0 & 0
\end{pmatrix} \in \mathfrak{g}$ as possible for which the symmetry condition \eqref{eq:checkAffine} (or \eqref{eq:checkAffineLd} in case of discrete Lagrangians) holds and such that all $v^1,\ldots,v^{(K)}$ are linearly independent. Then (under non-triviality conditions on $\mathcal{L}$) the $K$ quantities provided by Equation \eqref{eq:AffConserved} or Equation \eqref{eq:AffConservedLd} are functionally independent and are conserved under motions.

%%%%%%%%%%%%%%%%%%%%%%%%%%%%%%%%%%%%%%%%%%%%%%%%%%%%%%%%%%%%%%%%%%%%%%%%%%%

\section{Proposed approach}
%Going point by point through our novelties and explaining them
Our scheme aims to obtain a model for the true Lagrangian of the system with no restriction on the structure of the Lagrangian function apart from its regularity. This is done by first learning a discrete Lagrangian function $\mathcal{\mathcal{L}}^{NN}_d$ parameterised by a fully connected multi-layer perceptron (MLP) corresponding to the discrete inverse modified Lagrangian $\mathcal{L}^{invmod}_d$. After the learning stage is completed, these functions are used for forward simulation with a variational integrator to obtain an accurate discretisation $q_k$ of the true dynamics. Then the continuous inverse modified Lagrangian is obtained using 
 $\mathcal{L}^{invmod}(q,\dot{q}) = \mathcal{\mathcal{L}}^{NN}_d(q-\Delta t  \dot{q}/2, q+\Delta t \dot{q}/2 )$ based on the variational midpoint rule and employing the theory of VBEA from Section \ref{section:BEA}, an expression of the true Lagrangian $\mathcal{L}^{VBEA}$ is obtained and used to identify the corresponding Hamiltonian $\mathcal{H}^{VBEA}$.
 
To learn $\mathcal{L}^{NN}_d$ we rely on observations and simulations only of the configuration in time and do not pose any restrictions on its form. If required, velocity data at step $k$ can be computed using central difference or by solving
\[
-D_1 \mathcal{L}^{NN}_d(q_k,q_{k+1})= \nabla_{\dot q} \mathcal{L}^{VBEA}(q,\dot{q})   |_{(q_k , \,\dot{q}_k)}
\]
for $\dot{q}_k$. Observations and predictions are created from configuration data using three consecutive values in discrete time $(q_{k-1},q_{k},q_{k+1})$ from the same test trajectory. Each triple is used to compute  $D_2 \L^{NN}_d(q_{k-1},q_k) + D_1 \L^{NN}_{d}(q_{k},q_{k+1})$. As known from Equation (\ref{eqn:cELE}) these values should be zero for all discrete points of the trajectory. As there is no requirement on the sequentiality between triple observations the neural network can be trained by minimising 
\begin{equation}
    \min \frac{1}{N} \sum_{k=0}^{N-1}\sum_{i=1}^{n_q}(D_2 \L^{NN}_d(q_{k-1},q_k) + D_1 \L^{NN}_{d}(q_{k},q_{k+1}))[i])^2
    \label{eq:naive-cost-function} 
\end{equation} 
via a (stochastic) gradient descent algorithm and backpropagation to compute the derivative of the loss function with respect to the parameters (here $[\cdot]_i$ denotes the $i$-th element of the vector). % for all $1<k<N-1$. 

However, using only Equation \eqref{eq:naive-cost-function} without properly safeguarding the learning process can learn a model in which the roots of the equations of motion (the trajectory values) are degenerate, only touching the zero axis and not necessarily crossing it. Such roots present difficulties for forward simulation as they are difficult to find using common numerical methods such as Newton-Rhapson. To prevent such behaviour we propose adding the \textit{degeneracy} term 
   \begin{align}
       \frac{1}{N} \sum_{k=0}^{N-1}  \Big(1- \frac{1}{1+e^{-0.01(d_k)^m}}\Big), \notag \\
        d_k =  \textrm{det}\Big(D_2 D_1\L^{NN}_{d}(q_{k},q_{k+1})\Big ) \notag
    \end{align}
to the cost function \eqref{eq:naive-cost-function}, where $m=1 \textrm{ or } 2$ and $d$ corresponds to the Jacobian of the discrete equations of motion (\ref{eqn:dELE}).
This term aims to increase the slope of the equations of motion at the root locations, safeguarding against degenerate roots (which have vanishing gradients at the root location) and rewarding steeper crossings. This term also prevents the learning of constant discrete Lagrangian functions, which satisfy the equations of motion but do not result in meaningful dynamics.
    
    \subsection{Incorporation of symmetry into the loss function}
Using the proposed loss function we learn a non-degenerate discrete inverse modified Lagrangian, which can then be used to predict motions. However, since discrete Lagrangians are not uniquely determined by the system's motion, the learned discrete Lagrangian for a dynamical system which exhibits symmetries can be arbitrarily unsymmetric even if the learnt discrete Lagrangian minimises the loss function. 
In the following, we introduce a method which automatically detects symmetries of the dynamical system and drives the learnt discrete Lagrangian towards a symmetric representation. This is useful for system identification as symmetries inform us about conserved quantities which constitute important qualitative features of the dynamical system. Moreover, driving the learnt Lagrangian to a symmetric representation acts as an additional regulariser in the learning process and can be beneficial for numerical simulations of the learned system.

Based on the symmetry condition \eqref{eq:checkAffineLd} of Section \ref{sec:LieGroup}, to detect an affine linear symmetry of the dynamical system we propose adding the following term to the cost function:
\begin{align*}\label{eq:ellsym}
\ell_{\mathrm{sym}} = 
\frac 1N  \sum\nolimits_{k=0}^{N-1}  |(Mq_k+w)^\top \nabla_{q_k} \mathcal{L}_d^{NN} (q_k,q_{k+1})\\
    +(Mq_{k+1}+w)^\top \nabla_{q_{k+1}} \mathcal{L}_d^{NN} (q_k,q_{k+1})|^2,
\end{align*}
where $q_0,\ldots,q_N$ is a training trajectory. Additionally, the above expression is summed over all training trajectories.
The $n_q^2$ elements in the matrix $M$ and the $n_q$ elements in $w$ are now trainable parameters in addition to the parameters of $\mathcal{L}_d^{NN}$ and are included in the minimisation process. Furthermore, we add the non-triviality condition \begin{equation}\label{eq:nontrivsym}|\| M\|^2 + \|w\|^2 -1|^2 = 0\end{equation} to the loss function to encourage the process to learn a non-trivial symmetry. After learning, Equation~\eqref{eq:AffConservedLd} is a candidate for a conserved quantity of the discrete system and Equation \eqref{eq:AffConserved} of the underlying continuous system.

\begin{rem}
The choice to test the symmetry condition \eqref{eq:checkAffineLd} on points $( q_k, q_{k+1})$ in the training data to construct $\ell_{\mathrm{sym}}$ is arbitrary. In principle, $\ell_{\mathrm{sym}}$ could be any approximation (e.g.\ a Monte-Carlo integration) of 
\begin{align*}
\frac{1}{ \mathrm{vol}(\mathcal{M}^{\mathrm{o}}) } \int_{\mathcal{M}^{\mathrm{o}}} |(M q_0+w)^\top \nabla_{q_0} \mathcal{L}_d^{NN} ( q_0, q_{1})\\
    +(M q_{1}+w)^\top \nabla_{q_{1}} \mathcal{L}_d^{NN} ( q_0, q_{1})|^2 \d  q_0 \d  q_1
\end{align*}
where $\mathcal{M}^{\mathrm{o}} \subset Q \times Q$ is a (topologically open, pre-compact) subset of the discrete phase space covering all parts of interest.
\end{rem}

\begin{rem}
    To learn $K$ functionally independent integrals of motions, we simply add $K$ instances $\ell_{\mathrm{sym}}^{(j)}$ of $\ell_{\mathrm{sym}}$ from Equation \eqref{eq:ellsym} to the loss function. Each $\ell_{\mathrm{sym}}^{(j)}$ has trainable parameters $(M^{(j)},w^{(j)})$. Further, the non-triviality condition \eqref{eq:nontrivsym} is added for each instance of $\ell_{\mathrm{sym}}^{(j)}$. To make sure that all learned $(M^{(k)},w^{(k)})$ yield functionally independent integrals of motions, we need to make sure that they span a basis of the Lie algebra \eqref{eq:gAffSym}. This is achieved by adding the orthogonality condition 
    \[
    \sum\nolimits_{k=2}^K\sum\nolimits_{s=1}^{k-1}( \mathrm{vec}(M^{(s)})^\top \mathrm{vec}(M^{(k)}) + (w^{(s)})^\top w^{(k)})
    \]
    to the loss function. Here $\mathrm{vec}$ writes the columns of a matrix into a single vector for the computation of the Frobenius inner product of two matrices. This corresponds to learning a $K$-dimensional symmetry group.
\end{rem}

\begin{rem}
    The approach can be generalised to arbitrary Lie group actions as considered at the end of Section \ref{sec:LieGroup} as follows.
    For $k=1,\ldots, K$ define $\ell_{\mathrm{sym}}^{(k)}$ as a numerical approximation of
\begin{equation}\label{eq:ellsym}
\ell_{\mathrm{sym}}^{(k)} \approx \frac{1}{ \d \mathrm{vol}(\mathcal{M}^{\mathrm{o}}) } \int_{\mathcal{M}^{\mathrm{o}}} | \hat{v}^{(k)}(\mathcal L_d)|^2 \d  q_0 \d  q_1
\end{equation}
%Here $\d \mathrm{vol}$ is a volume form on $Q \times Q$.
Here $\hat{v}^{(k)}$ denotes the invariant vector field to $v$.
The term $\ell_{\mathrm{sym}}^{(k)}$ measures how invariant $\mathcal L$ is under actions with group elements of $\exp({t v^{(k)} } | t \in \R)$.
Equip $\mathfrak{g}$ with an inner product $\langle \cdot , \cdot \rangle $ and norm $\| \cdot \| $.
Given weights $\alpha^{(k)},\beta^{(k)} >0$ define
\begin{align}\label{eq:l_sym}
\ell^{\mathrm{total}}_{\mathrm{sym}} = \sum_{k=1}^K \left(\ell_{\mathrm{sym}}^{(k)}+ \alpha^{(k)} | \| v^{(k)}\| -1 |^2
+ \beta^{(k)} \sum_{s=1}^{k-1} \langle v^{(k)},v^{(s)} \rangle \right) \notag
\end{align}
which is added to the loss function. Here $\alpha^{(k)}$ and $\beta^{(k)}$ are fixed, non-negative weights. 
The last two terms of $\ell_{\mathrm{sym}}$ measure the orthonormality of the spanning set $v^{(1)},\ldots,v^{(K)}$ while the first term measures how well infinitesimal actions by elements of $\exp(V)$ preserve $\mathcal L$.
\end{rem}

%%%%%%%%%%%%%%%%%%%%%%%%%%%%%%%%%%%%%%%%%%%%%%%%%%%%%%%%%%%%%%%%%%%%%%%
\section{Results}

To evaluate the proposed DLNN and SymDLNN methods alongside LNN, each of the methods was applied to two systems: a pendulum on a cart defined by the Lagrangian in Equation (\ref{eqn:LCart}) with $m_1 =m_2=l=1$, and the translational symmetry with $M=[0,0;0,0]$, $w=[1,0]$
and the Kepler problem in two dimensions with Lagrangian 
\begin{equation}
    \mathcal{L}_{\mathrm{Kp}}(x,y,\dot{x},\dot{y})= \frac{1}{2}(\dot{x}^2+\dot{y}^2)+ \frac{Gm_1m_2}{\sqrt{x^2 + y^2}}
\end{equation}
for $q = [x,y]^T$, $G=6.673 \times 10^{-26}$, $m_1=6\times 10^{24}$, $m_2=100$
and rotational symmetry of the form $w=[0,0]$, $M=[0,\sqrt{2}/2;-\sqrt{2}/2,0]$. 
For each system, the methods are tasked with learning the underlying dynamics defined by the Lagrangian in three evaluation scenarios: a single trajectory case, a multi-trajectory case, and learning a single trajectory with noisy measurements. After training a variational integrator is used for simulations with each of the three methods in order to assess how well the resulting system model is capable of: recreating the trajectory it was trained on, its ability to predict and extension of this trajectory and ability to preserve the energy and symmetry. Code for all implementations and experiments can be found at https://github.com/yanalish/SymDLNN.

\begin{figure} [!tbp]
    \centering
  \includegraphics[scale=0.55]{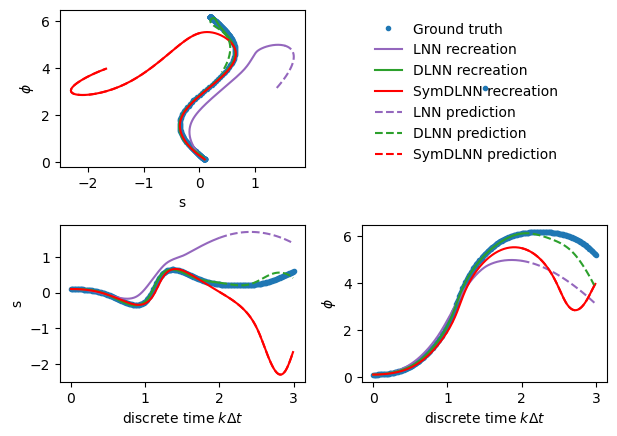}
    \caption{Pendulum on the cart example: Recreation and prediction of the trajectory based on single trajectory observation.}
    \label{fig:trajplotcartpend}
\end{figure}

\begin{figure} [!tbp]
    \centering
  \includegraphics[scale=0.55]{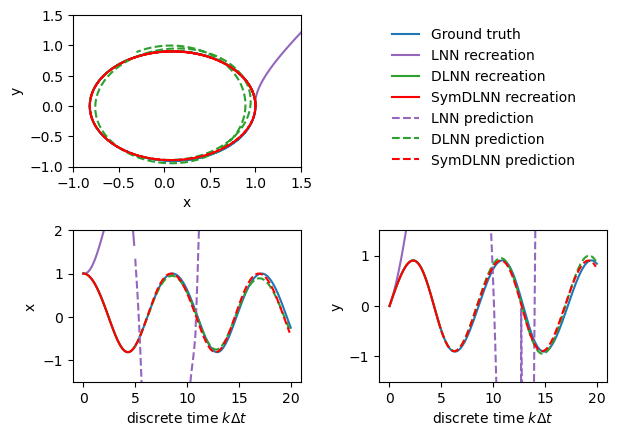}
    \caption{ Kepler example: Recreation and prediction of the trajectory based on single trajectory observation.}
    \label{fig:trajplotKepler}
\end{figure}

In the first scenario a set of experiments are conducted wherein the observations used to train the neural network are obtained from a single trajectory of $N$ consecutive configuration points spaced at a stepsize of $\Delta t$. For the pendulum on a cart $N=200$ with $\Delta t=0.01$ was chosen, whereas for the Kepler problem these were set to $N=50$ and $\Delta t=0.1$  Trajectory observations were obtained using a variational integrator with a fine grained step size $\Delta t = 10^{-4}$ for the pendulum on a cart and $\Delta t = 10^{-3}$ for the Kepler problem. Using these observations each method attempts to recreate the trajectory with the learnt Lagrangian and predict its extension for an additional $N_{extra} = 100$ steps for the pendulum and $N_{extra} = 150$ for the Kepler problem, the resulting trajectory will be denoted as $q^{NN}$. 
For a fair comparison each of LNN, DLNN and SymDLNN methods approximate the continuous and discrete Lagrangian respectively using a 3-layer multi-layer perceptron (MLP) with 128 dimensional hidden layers and SoftPlus activations at each node. The minimisation of the objective functions is performed through gradient descent. Specifically, for all methods the optimisation is performed using an Adam optimiser with an initial learning rate of $3 \times 10^{-3}$, other Adam hyperparameters were set to $\beta_1 = 0.9$, $\beta_2 = 0.999$, and $\epsilon = 1\times10^{-8}$. Each network was trained for 100,000 epochs. The only notable difference in the training between the methods is in the SymDLNN method which does not employ the symmetry term for the first 5,000 epochs and then utilises them in the remaining 95,000 epochs. The rationale for this is to initially optimise the parameters of the network to find a suitable initial Lagrangian as the DLNN does and then discover the symmetries.

\begin{figure} [!tbp]
    \centering
  \includegraphics[scale=0.55]{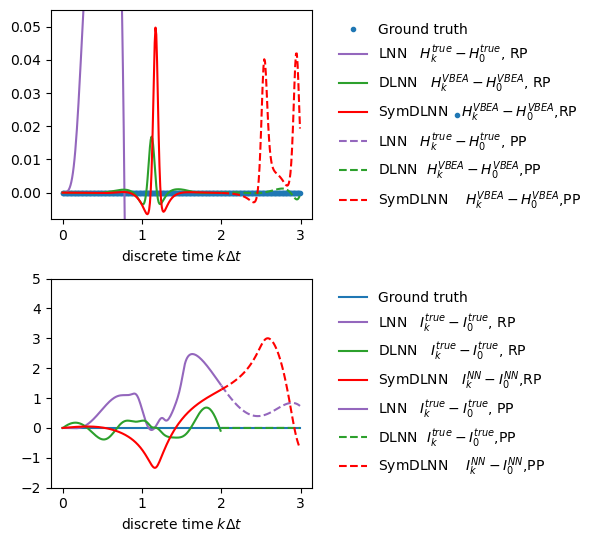}
    \caption{Pendulum on a cart example: Symmetry and energy error test based on single trajectory observation (RP- recreation phase, PP- prediction phase).}
    \label{fig:singletraj_HIpred_A}
\end{figure}

\begin{figure} [!tbp]
    \centering
  \includegraphics[scale=0.55]{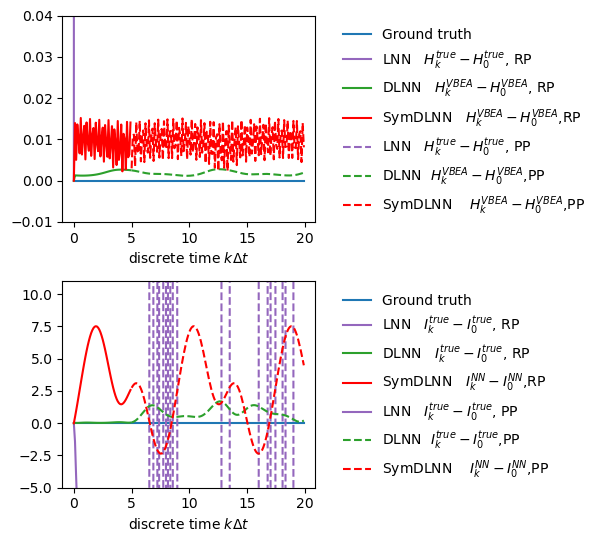}
    \caption{Kepler example: Symmetry and energy error test based on single trajectory observation (RP- recreation phase, PP- prediction phase).}
    \label{fig:singletraj_HIpred_C}
\end{figure}

In Figures \ref{fig:trajplotcartpend} and \ref{fig:trajplotKepler} we can see for both example systems that the DLNN and SymDLNN methods better recreate the original train trajectory than the LNN approach. For the pendulum on a cart example DLNN provides the best prediction whereas for Kepler both DLNN and SymDLNN outperform LNN for the unseen extension. For the cart pendulum example the SymDLNN model identified symmetry parameters $M^{NN}_{cp}=[ 1.51\times 10^{-3}, -1.26\times 10^{-3}; -7.83\times 10^{-6}, -3.62\times 10^{-5}]$, $w^{NN}_{cp} = [9.97\times 10^{-1},  8.04\times 10^{-2}]^T$ and for the Kepler $M^{NN}_{Kp}=[ 0.074,  0.813, -0.568, -0.018]$, $w^{NN}_{Kp} = [0.057, 0.076]^T$ from initialization guess $M^{guess}_{cp}=[0.1, 0.1 ;  0.1, 0.1]$, $w^{guess}_{cp} = [1.5,0.5]^T$ and $M^{guess}_{Kp}=[0.1, 0.807 ;  -0.607, 0.1]$, $w^{guess}_{Kp} = [0.1,0.1]^T$ respectively. We believe this can be further improved with a more rigorous hyperparameter search and more generous computing resources. Figures \ref{fig:singletraj_HIpred_A} and \ref{fig:singletraj_HIpred_C} demonstrate how accurately the symmetry and energy are preserved with each model. In these plots, we compare how well LNN and DLNN preserve the symmetry with
\begin{align}
    &I^{true}_k = (p_k^{NN})^T(Mq_k^{NN}+w) \notag \\ 
    \textrm{for } & p_k^{NN} = \nabla_{\dot{q}} \mathcal{L}^{true}\Big(q_k,\tfrac{q_{k+1}^{NN}-q_{k-1}^{NN}}{2\Delta t}\Big) \notag
    \end{align}
and how well SymDLLN learns it by computing
\begin{align}
    &I^{NN}_k = (p_k^{NN})^T(M^{NN}q_k^{NN}+w^{NN}) \notag \\  \textrm{for } 
   & p_k^{NN} = -D_1 \mathcal{L}^{NN}_d(q^{NN}_k,q^{NN}_{k+1})\notag
    \end{align}
For the energy plots we compare how well LNN has learnt the original Hamiltonian and how accurately a Hamiltonian has been approximated based on VBEA using the DLNN and SymDLNN model calculating respectively $H^{true}_k(q_k,\tfrac{q_{k+1}^{NN}-q_{k-1}^{NN}}{2\Delta t})$ and $H^{VBEA}_k(q_k,\tfrac{q_{k+1}^{NN}-q_{k-1}^{NN}}{2\Delta t})$. 
We can clearly see the improvement of using DLNN compared to the LNN approach. The SymDLNN approach performs slightly worse, however, it is important to note that for LNN and DLNN we plot based on our knowledge for the symmetry expression whereas SymDLNN has learnt the symmetry itself, providing a qualitatively better model and additional knowledge of the system's dynamics. 

 Both examples represent non-chaotic systems, thus learning from a single trajectory learns the behaviour of the system for a restricted portion of the phase space. For this purpose, the same experiments were conducted for all three methods when learning from data from 100 different trajectories of the same lengths and steps as in the single trajectory examples. Recreation and prediction of the trajectory as well as symmetry and energy preservation tests closely resembled the results demonstrated above for learning from a single trajectory. 

\begin{figure} [!tbp]
    \centering
  \includegraphics[scale=0.57]{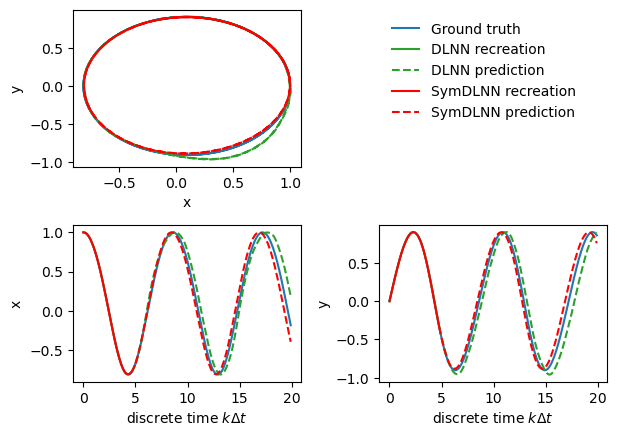}
    \caption{Kepler example with noise: Recreation and prediction of the trajectory based on observations gathered from a single trajectory.}
    \label{fig:trajplot4}
\end{figure}

\begin{figure} [!tbp]
    \centering
  \includegraphics[scale=0.55]{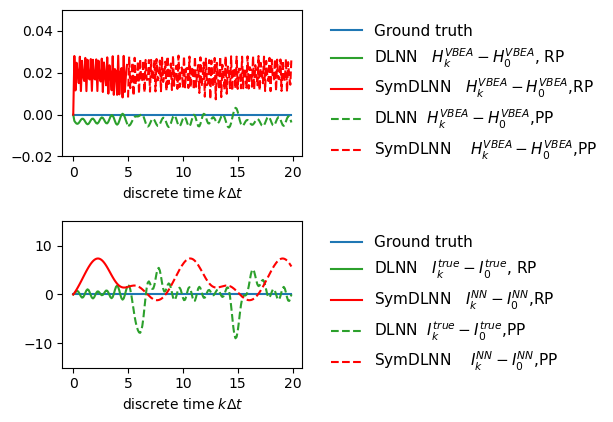}
    \caption{Kepler example with noise: Symmetry and energy error test based on single trajectory observation (RP- recregation phase, PP- prediction phase).}
    \label{fig:singletraj_HIpred_E}
\end{figure}

In our third and last evaluation scenario we examined the performance of our approaches DLNN and SymDLNN in the presence of noise. As we can see below for the Kepler problem the proposed model was capable of accurately recreating the observed trajectory and extending it beyond for $N= 50$ and $N_{extra}=200$ learning from a single trajectory observation for which measurement noise was added with a standard Gaussian distribution with $\sigma^2=0.001$. For DLNN, the $H^{VBEA}$ energy and the true symmetry are quite well preserved both during the recreation and the prediction of the trajectory. Despite the noise, SymDLNN is capable of preserving energy and although the symmetry oscillation is slightly larger, it is based on a symmetry which the SymDLNN approach itself identified with values \textbf{ $w^{NN} = [ 0.056,  0.077]^T , M^{NN} =  [0.090,  0.812, -0.568, -0.016]$} with the same initialization as before $M_{Kp}^{guess}$, $w_{Kp}^{guess}$. Moreover, we can see that the additional symmetry term helps SymDLNN better extend the trajectory than DLNN.

\section{Conclusion}

We have presented two novel methods, DLNN and SymDLNN, which successfully incorporate the use of variational integrators and variational backward error analysis for learning discrete Lagrangians, utilising the underlying structure of the problem to improve the learned system representation qualitatively and guard against numerical issues in forward simulations. This was achieved through the proposal of a novel regularisation degeneracy term that inductively biases the network to avoid learning Lagrangians whose equations of motion have degenerate roots as well as constant Lagrangians. SymDLNN further extends the DLNN in its ability to automatically identify symmetries during the learning process through a framework that identifies a subgroup which acts by symmetries upon a given Lie group on the configuration space. Numerical experiments demonstrate the qualitative improvements of our proposed methods to previous approaches.
% is this double blind?

\begin{ack}
Y.L. acknowledges funding from the EPSRC Doctoral Training Partnership EP/R513295/1, project reference 2280382. P.S. acknowledges funding by the W.D. Armstrong Fund from the School of Technology at the University of Cambridge. 
S.R. acknowledges funding by the EPSRC Centre for Doctoral Training in Autonomous Intelligent Machines \& Systems
EP/L015897/1. C.O.\ acknowledges funding by the Ministry of Culture and Science of the State of North Rhine-Westphalia.
\end{ack}
\bibliography{ifacconf} 

\end{document}